\documentclass[journal]{IEEEtran}

\ifCLASSINFOpdf
\else
\fi
\usepackage[colorlinks,urlcolor=blue,linkcolor=blue,citecolor=blue]{hyperref}
\usepackage{dblfloatfix}
\usepackage{graphicx}
\usepackage{csquotes}
\usepackage{multirow}
\usepackage{mathtools}
\usepackage{amsmath}
\usepackage{amssymb}
\usepackage{url}
\usepackage{makecell}
\usepackage{graphicx}
\usepackage{hyperref}
\usepackage{epstopdf}
\usepackage{algorithm}
\usepackage{subcaption}
\usepackage{epstopdf}
\usepackage[nocompress]{cite}
\usepackage{xcolor}
\usepackage{mathtools}
\usepackage{ragged2e}
\usepackage{color, colortbl}
\definecolor{Gray}{gray}{0.9}

\begin{document}

\title{HateTinyLLM : Hate Speech Detection Using Tiny Large Language Models}

\author{Tanmay Sen, Ansuman Das,  Mrinmay Sen
\IEEEcompsocitemizethanks{\IEEEcompsocthanksitem  Tanmay Sen is with Ericsson, Kolkata, e-mail:sentanmay518@gmail.com.
\IEEEcompsocthanksitem Ansuman Das is with Ericsson, Kolkata,e-mail:ansumandasiiit@gmail.com.
\IEEEcompsocthanksitem Mrinmay Sen is a joint research scholar in the Department of Artificial Intelligence, Indian Institute of Technology Hyderabad and the Department of Computing Technologies, Swinburne University of
Technology, Australia, e-mail: msen@swin.edu.au}}
\maketitle

\begin{abstract}
Hate speech encompasses verbal, written, or behavioral communication that targets derogatory or discriminatory language against individuals or groups based on sensitive characteristics. %The applications of hate speech detection includes employing content moderation algorithms and human moderators to detect and remove hate speech from online platforms, implementing legal frameworks to enforce consequences for perpetrators, utilizing social media monitoring tools to analyze trends and patterns, and conducting educational campaigns to raise awareness and foster digital literacy. 
Automated hate speech detection plays a crucial role in curbing its propagation, especially across social media platforms. Various methods, including recent advancements in deep learning, have been devised to address this challenge. In this study, we introduce HateTinyLLM, a novel framework based on fine-tuned decoder-only tiny large language models (tinyLLMs) for efficient hate speech detection. Our experimental findings demonstrate that the fine-tuned HateTinyLLM outperforms the pretrained mixtral-7b model by a significant margin. We explored various tiny LLMs, including PY007/TinyLlama-1.1B-step-50K-105b, Microsoft/phi-2, and facebook/opt-1.3b, and fine-tuned them using LoRA and adapter methods. Our observations indicate that all LoRA-based fine-tuned models achieved over 80\% accuracy.
\end{abstract}


\section{Introduction}
\IEEEPARstart{H}ate speech detection \cite{davidson2017automated, fortuna2018survey, schmidt2017survey} in text data has garnered significant attention in recent years, with researchers exploring various approaches to address this complex problem. The task of hate speech detection refers to identifying and categorizing language that expresses hatred, prejudice, or hostility towards individuals or groups based on attributes such as race, ethnicity, religion, gender, sexual orientation, disability, or other protected characteristics. The goal of hate speech detection is to develop automated systems or algorithms that can analyze text data, such as social media posts, comments, or news articles, and identify instances of hate speech. Efficiently detecting and mitigating hate speech can help to protect individuals and communities from the negative consequences such as discrimination, violence and social division. Various methodologies and  data sets have been explored and generated for hate speech detection problem.  All the previously proposed methodologies can be broadly categorized into  three groups: traditional machine learning, deep learning methods that utilized word embedding and transformers-based encoders only methods.  Malik et al. \cite{malik2022deep} present a comparative study on fourteen different deep learning models, concluding that transformers-based hate speech detection models exhibit more promising results than classical and embedding based deep learning models. Various deep learning models like as LSTM, biLSTM, and convolution neural network with Word2Vec embedding are employed by various researchers \cite{zhang2018detecting} for hate speech detection. \cite{badjatiya2017deep} conduct comprehensive experiments utilizing various deep learning models to acquire semantic word embeddings.

 Transformer models \cite{vaswani2017attention}  ,  like BERT \cite{devlin2018bert}, ELECTRA \cite{clark2020electra}, and BART \cite{lewis2019bart} offer superior syntactic and semantic understanding of words within text compared to traditional word2vec or GloVe vectors for word embedding. Mozafari et. al. \cite{mozafari2020bert} explore BERT's capability to capture hateful context within social media content using novel fine-tuning methods. A fine-tune multilingual BERT model  for hate speech detection is proposed by  Aluru et al.\cite{aluru2021deep} in low-resource languages. Graph neural network (GNN) based mutlitask framework MTBullyGNN for cyberbullying detection is proposed by Maity et al.  \cite{maity2022mtbullygnn} for  code-mixed language. They utilized a sentence encoder and BART model to represent text. Awal et. al.  \cite{awal2023model} propose HateMAML, a novel meta learning based approach for hate speech detection in low-resource languages. 
% Hate speech detection often involves natural language processing (NLP) techniques, machine learning models, and sometimes deep learning approaches. These systems analyze textual features, linguistic patterns, and contextual information to distinguish between hateful and non-hateful language. However, it's a challenging task due to the complexity and subjectivity of language, cultural nuances, evolving forms of hate speech, and the need to balance free speech with the prevention of harm.
Recently, researchers have explored various techniques to enhance the performance and robustness of large language models (LLM) based hate speech detection. Guo et al. \cite{guo2023investigation} conduct large scale study on effective prompting methods for context-aware LLM-based hate speech detection to improve accuracy. Often, hate speech detection problem suffers lack of label data,  Pendze et al. \cite{pendzel2023generative} mitigate this issues by generating large amount of synthetic data using LLM. The authors compare two types of enocoder and decoder only transformer models.  They have observed zero shot GPT3.5 performs marginally better than fine tuned BERT family of models.   

Tiny LLMs are advantageous over large LLMs due to their reduced computational requirements, faster inference times, and lower memory footprint. Therefore, tiny LLMs are often more suitable for niche applications where a smaller model may suffice without sacrificing much performance. This work introduces the use of various fine-tuned decoder only based tiny large language models (LLMs) for hate speech detection. These tiny LLMs offer strong performance across various tasks while requiring fewer resources, making them ideal for practical applications. Notably, this is the first study to explore leveraging fine-tuned decoder only tiny LLMs for hate speech detection.

\begin{table}[htbp]

\centering
\caption{Samples from DynaHate dataset}
\label{dynahate}
\begin{tabular}{p{5cm} c}
\hline
\textbf{Tweet} & \textbf{Hate Class} \\
\hline
dalits are lowlives & hate \\
%\hline
gay people are a burden to our society & hate \\
%\hline
Arabs are not welcome & hate \\
%\hline
i wish corona didnt break this country & nothate \\
%\hline
asian people are the shit & nothate \\
%\hline
i hate wearing black in the summer! & nothate \\
\hline
\end{tabular}
\label{tab:tweets}
\end{table}

\begin{table}[htbp]
\centering
\caption{Samples from Hateeval dataset}
\label{hateeval}
\begin{tabular}{p{5cm}c}
\hline
\textbf{Tweet} & \textbf{Hate Class} \\
\hline
White women are stupid & hate \\
%\hline
Any woman that likes me is a bitch. All women are bitches. & hate \\
%\hline
Shut the fuck up pls & hate \\
%\hline
It must be amazing to be me & nothate \\
%\hline
Only I could get kicked out of a club for just being & nothate \\
%\hline
They're replaceable, your time is not & nothate \\
\hline
\end{tabular}
\label{tab:tweets}
\end{table}

\section{Datasets}
To begin, we reviewed the literature for the existing hatespeech detection datasets in English. We have considered two hatespeech datasets ({\em DynaHate} \cite{vidgen2020learning} and {\em hateeval} \cite{basile2019semeval}) in English for our experiment. DynaHate dataset comprises approximately 41,144 entries, as shown in Table~\ref{datasetDistribution}. This dataset is created through a collaborative human-and-model-in-the-loop process aimed at enhancing hate detection models. This approach facilitated the collection of four rounds of datasets specifically focused on hate speech. In the balanced DynaHate dataset, comprising 411,144 entries, tweets are evenly distributed with 46\% percent classified as 'Not Hate' and 54\%  as 'Hate,' ensuring robust representation across categories. The HateEval dataset looks at hate speech aimed at women and immigrants on Twitter. It has about 9000 entries. Among these, 58\% of the tweets are not hateful, while 42\% contain hate speech. 
%Unlike other dataset, HateEval doesn't have an equal number of both types of tweets. 
Some samples of both the {\em DynaHate} and {\em Hateeval} dataset are shown in Table \ref{dynahate} and Table  \ref{hateeval} respectively. Detailed class-wise distributions of {\em DynaHate} and {\em Hateeval} datasets are also given in Table \ref{datasetDistribution}. 

% \begin{table}[h]
% \centering
% \caption{Dataset Summary}
% \begin{tabular}{cccccc}
% \hline
% \multirow{2}{*}{\textbf{Class}} & \multicolumn{2}{c}{\textbf{DynaHate}} & \multicolumn{2}{c}{\textbf{Hateeval}} \\
% \cline{2-5} 
%  & \textbf{Tweet}    &   \textbf{Tweet}  \\
% \hline
% \textbf{Hate} & 22175 & 3783  \\
% \textbf{NotHate} & 18969 & 5217  \\
% \hline
% \end{tabular}
% \end{table}

\begin{table}[h]
\centering
\caption{Dataset Summary}
\label{datasetDistribution}
\begin{tabular}{ccc}
\hline
\textbf{Class} & \textbf{DynaHate} & \textbf{HateEval} \\
\hline
\textbf{Hate} & 22175 & 3783 \\
\textbf{NotHate} & 18969 & 5217 \\
\hline
\end{tabular}
\end{table}

\section{Methodology}
In the following section dives into formulating the problem and unveils a framework for hate speech detection with various tiny LLMs. Zhang et al. 
Zhang et al. \cite{zhang2024tinyllama} introduces TinyLlama a condensed language model comprising 1.1 billion parameters. It trained on approximately 3 trillion tokens across three epochs. TinyLlama extends the architecture and tokenizer initially developed for Llama 2.
Employing advancements such as flashAttention, TinyLlama achieves superior computational efficiency and exhibits impressive performance across various downstream tasks, surpassing existing open-source models of comparable sizes. By training smaller models with larger datasets, the study explores the potential of optimizing performance within specific inference constraints, challenging the preference for larger models. The pretraining process effectively combines natural language and code data, resulting in competitive performance. Through extensive experimentation and optimization, including speed enhancements like Fully Sharded Data Parallelism and flash attention, TinyLlama showcases superior training efficiency and problem-solving capabilities. The paper underscores the significance of smaller, efficient models like TinyLlama in enhancing accessibility and promoting innovative research in language model development.TinyLlama consists of 22 layers,  16 attention heads, with an embedding size of 2048.

Li et al. \cite{li2023textbooks} propose Phi, represents a significant advancement in the realm of smaller-scale transformers, demonstrating impressive performance across various benchmarks without the need for an extensive parameter count. Its utilization of a diverse range of data sources, including synthetic texts and filtered websites, highlights a strategic approach to training that enriches its understanding of language and common sense. Notably, Phi-2's ability to achieve near-state-of-the-art performance with just 2.7 billion parameters underscores the importance of efficient model design and data augmentation techniques. Moreover, its focus on safety and educational value, reflected in the careful curation of data sources, speaks to a conscientious approach to AI development.Phi model features 24 layers with 32 attention heads, each having a dimension of 32. Its context length is set as 2048.

Zhang et al. \cite{zhang2023opt} present open pre-trained transformers (OPT), a collection of decoder-only pre-trained transformers with parameter sizes ranging from 125 million to 175 billion ( in our study we have used 1.3 billion parameters model) aiming to facilitate reproducible and responsible research in large language models (LLMs). The authors highlight the limited access to full model weights of existing LLMs and the significant computational cost involved in training such models. They present detailed architectural specifications and training methodologies, emphasizing transparency and efficiency. Evaluation results across various NLP tasks, dialogue datasets, bias, and toxicity benchmarks demonstrate the competitiveness of OPT-175B compared to existing models like GPT-3 Davinci and PaLM. While OPT-175B generally matches or outperforms existing models in NLP tasks and dialogue generation, it exhibits higher stereotypical biases and toxicity rates, indicating the need for further research on ethical considerations and model improvements.Opt 1.3B consists of 24 layers, each containing 32 attention heads, with an embedding size of 2048.

\begin{figure}[!h]
  \centering
  \includegraphics[width=1.5\linewidth]{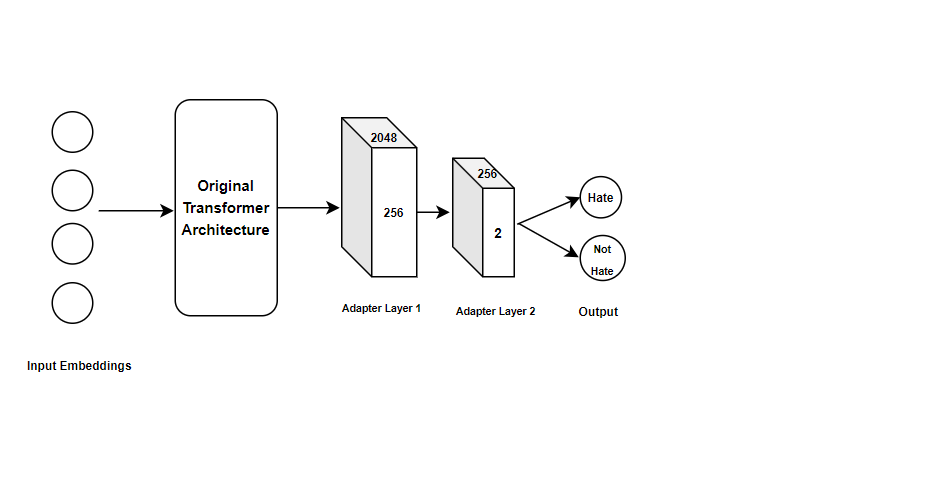}
  \caption{Adapter architecture }\label{fig:1}
\end{figure}

\begin{figure*}[!h]
  \centering
  \includegraphics[width=1\linewidth]{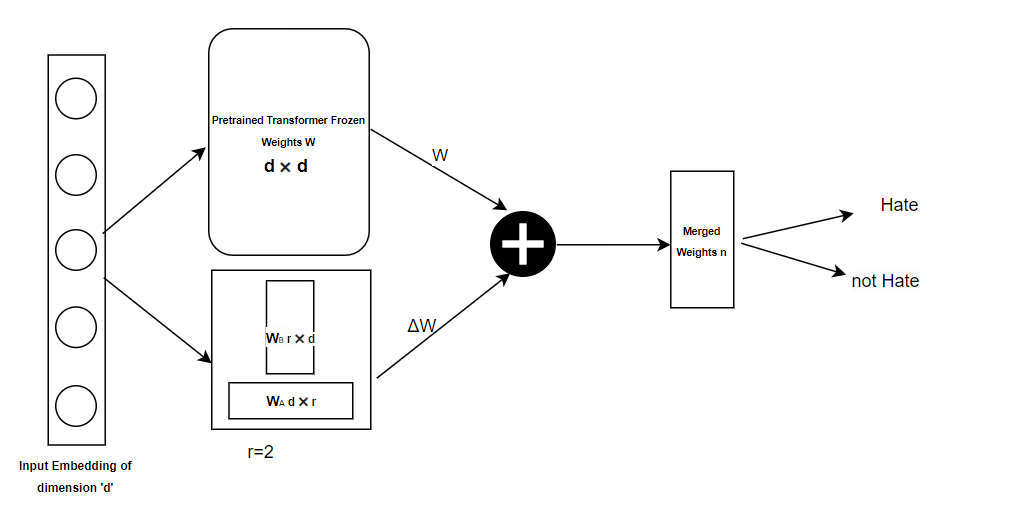}
  \caption{LoRA architecture }\label{fig:2}
\end{figure*}
We have used  Low-Rank Adaptation (LoRA) and Adapter methods for our small LLm fine-tuning purpose.
The paper \cite{hu2021LoRA} introduces LoRA as a solution to the challenge of fine-tuning large pre-trained models like GPT-3 (175 billion parameters) for specific tasks, which can be prohibitively expensive due to the sheer size of the model. The authors acknowledge the paradigm of pretraining on general domain data and adapting to particular tasks, but note that full fine-tuning becomes less feasible as models grow larger. To address this issue, the proposed LoRA approach involves freezing the weights of the pretrained model and introducing trainable rank decomposition matrices into each layer of the Transformer architecture. This effectively reduces the number of trainable parameters for downstream tasks while still allowing adaptation. The proposed architecture can be foiund in Fig\ref{fig:1}.

\begin{table}[!hbt]
\centering
\caption{Hyperparameters for LoRA}
\begin{tabular}{cccc}
\hline
\textbf{Hyperparameters} & \textbf{TinyLlama} & \textbf{phi-2} & \textbf{opt-1.3B} \\ \hline
Epoch & 3 & 3 & 3 \\ \hline

Target Modules & \shortstack{k\_proj',\\'v\_proj} & \shortstack{k\_proj',\\'v\_proj} & \shortstack{k\_proj',\\'v\_proj} \\ \hline
\ Trainable Parameters & 0.01 \%\ & 0.02 \%\ & 0.03\%\ \\ \hline
LoRA\_alpha & 16 & 16 & 16 \\ \hline
r & 2 & 2 & 2 \\ \hline
LoRA\_dropout & 0.05 & 0.05 & 0.05 \\ \hline
Batch Size & 8 & 8 & 8 \\ \hline
weight\_decay & 0.001 & 0.001 & 0.001 \\ \hline

Training Time & \shortstack{1.05 hour} & \shortstack{1.20 hour} & \shortstack{1.10 hour} \\ \hline
\end{tabular}
\label{tab:hyperparameters}
\end{table}

The adapter method, as introduced by Houlsby et al. (2019) \cite{houlsby2019parameter}, offers a parameter-efficient approach to enhancing large language models (LLMs) by adding adapter layers to transformer blocks. Unlike prefix tuning, which modifies embeddings, adapter layers are inserted into two positions within each transformer block. These adapter layers consist of relatively small fully connected layers with a bottleneck structure akin to autoencoders.  This design significantly reduces the number of parameters required compared to traditional methods
% \subsection{Problem Statement}

\section{Experiments, Results and Analysis}
This section describes the outcomes of raw pretrained models and our proposed finetuned models.

\subsection{Baselines Setup}
% The study began by employing the base model to classify tweets using zero-shot prompting across two datasets.
Utilizing all three Tiny LLMs, both before and after fine-tuning, as well as Mistral 7B,we have  evaluated the classification results. The experiments were conducted in a consistent computing environment. Notably, the baseline models demonstrated an average accuracy of 0.5, and no quantization was applied to these models.
All baseline model results are summarised in below tables.

% [{\includegraphics[width=4in,height=3.05in,clip{adapter.png}}]
% \includegraphics[width=4in,height=3.05in,clip]
% {adapter.png}
\subsection{Experimental Setup and Hyperparameters}

The experiments were conducted sequentially, with each method executed separately by restarting the kernel to ensure independent runs. A Nvidia P100 GPU with 16 GB memory was utilized for all experiments. In bothe the methods we have chnaged the weights for k-proj and v-proj layer fo the all 3 tiny llms.TinyLlama, phi-2, and opt-1.3B. Across all methods, epochs are set to 3, target modules focus on $k_proj$ and $v_proj$, and there's a slight variation in the percentage of trainable parameters. Additionally, common values for $LoRA_alpha$, r, $LoRA_dropout$, batch size, weight decay, and training time are maintained. In contrast, the second table presents the parameters for Adapter (LLM) and LoRA methods. While epochs are increased to 5 for both, training times differ slightly The fine-tuning process involved 5 epochs for the Adapter-based method and 3 epochs for the LoRA method, with detailed hyperparameters specified for each.
In both LoRA and adapter-based fine-tuning, the AdamW optimizer is utilized. Additionally, in the adapter-based method, the negative-log-likelihood loss function is employed as the loss function.

% \begin{table}[h]
%  \centering
%  \caption{Hyperparameters for Adapter}
%  \begin{tabular}{|c|c|c|c|}
%  \hline
%  \textbf{Adapter/LoRA Parameters} & \textbf{Epoch} & \textbf{Training Time} & \textbf{Adapter Layer Added} \\ \hline
%  TinyLlama & 5 & 1 Hour 5 Mins & 2 \\ \hline
%  phi-2 & 5 & 1 Hour 31 Mins & 2 \\ \hline
%  opt-1.3B & 5 & 1 Hour 5 Mins & 2 \\ \hline
%  \end{tabular}
%  \label{tab:adapter_q}
%  \end{table}

 % \begin{table}[h]
 % \centering
 % \caption{Hyperparamaters for Adapter }
 % \begin{tabular}{|c|c|c|c|}
 % \hline
 % \textbf{Adapter/LoRA Parameters} & \textbf{TinyLlama} & \textbf{phi-2} & \textbf{opt-1.3B} \\ \hline
 % Epoch & 5 & 5 & 5 \\ \hline
 % Training Time & 1 Hour 5 Mins & 1 Hour 31 Mins & 1 Hour 5 Mins \\ \hline
 % Adapter Layer Added & 2 & 2 & 2 \\ \hline
 % \end{tabular}

 % \label{tab:adapter_q}
 % \end{table}

\begin{table}[h]
\centering
\caption{Hyperparamaters for Adapter }
\begin{tabular}{cccc}
\hline
\textbf{Parameters} & \textbf{TinyLlama} & \textbf{phi-2} & \textbf{opt-1.3B} \\ \hline
Epoch & 5 & 5 & 5 \\ 
%\hline
Trainable Parameters & 0.05\% & 0.01\% & 0.03\%\\ 
%\hline
Adapter Layer Added & 2 & 2 & 2 \\
%\hline
Training Time & 1.05hour & 1.31 hour & 1.05 hour \\ 
 \hline
\end{tabular}
\label{tab:adapter_q}
\end{table}

 % \usepackage{multirow}
 % \usepackage{multirow}

\subsection{Results and Discussion}

In this work , we conducted a comparative analysis of four base models' performance on two distinct datasets: Dynahate and Hateeval with metrics including accuracy and F1 scores. From Table \ref{base_model}, Among the models assessed, TinyLlama demonstrated moderate performance, achieving an accuracy of 0.50 and an F1 score of 0.61 on DynaHate, while on Hateeval, its accuracy decreased to 0.29 with an F1 score of 0.24. phi-2 exhibited slightly better results, with an accuracy of 0.52 and an F1 score of 0.66 on DynaHate, and a corresponding accuracy of 0.47 and F1 score of 0.28 on Hateeval. opt-1.3b showcased comparable performance across both datasets, with an accuracy of 0.53 and an F1 score of 0.54 on DynaHate, and an accuracy of 0.45 with an F1 score of 0.17 on Hateeval. In contrast, Mistral-7B-v0.1 emerged as the top-performing model, with an accuracy of 0.58 and an F1 score of 0.52 on DynaHate, and notably higher scores on Hateeval, boasting an accuracy of 0.73 and an F1 score of 0.16. Overall, while some models displayed consistency across datasets, others demonstrated varying degrees of performance 
\begin{table}[hbt]
\centering
\caption{Base models performance on both the data sets}
\label{base_model}
\begin{tabular}{cccccc}
\hline
\multirow{2}{*}{\textbf{Model name}} & \multicolumn{2}{c}{\textbf{DynaHate}} & \multicolumn{2}{c}{\textbf{Hateeval}} \\
\cline{2-5} 
 & \textbf{Accuracy} & \textbf{F1} & \textbf{Accuracy} & \textbf{F1} \\
\hline
\shortstack{\textbf{TinyLlama}} & 0.50 & 0.61 & 0.29 & \textbf{0.24} \\
%\hline
\textbf{phi-2} & 0.52 & \textbf{0.66} & 0.47 & 0.28 \\
%\hline
\textbf{opt-1.3b} & 0.53 & 0.54 & 0.45 & 0.17 \\
%\hline
\textbf{Mistral-7B-v0.1} & \textbf{0.58} & 0.52 & \textbf{0.73} & 0.16 \\
\hline
\end{tabular}
\end{table}

 \begin{table}[hbt]
 \centering
 \caption{Adapter based finetuned models performance}
 \label{adapter}
 \begin{tabular}{ccccc}
 \hline
 \multirow{2}{*}{\textbf{Model name}} & \multicolumn{2}{c}{\textbf{DynaHate}} & \multicolumn{2}{c}{\textbf{Hateeval}} \\ \cline{2-5} 
  & \textbf{Accuracy} & \textbf{F1} & \textbf{Accuracy} & \textbf{F1} \\ \hline
 \shortstack{\textbf{TinyLlama-1.1B}} & 0.71 & 0.75 & 0.69 & 0.7 \\ %\hline
 \textbf{phi-2} & 0.7 & \textbf{0.76} & 0.72 & 0.71 \\ %\hline
 \textbf{opt-1.3b} & 0.71 & 0.71 & \textbf{0.72} & \textbf{0.74} \\ \hline
 \end{tabular}
 \end{table}

\begin{table}[hbt]
\centering
\caption{LoRA based finetuned models performance on both the data sets}
\label{lora}
\begin{tabular}{cccccc}
\hline
\multirow{2}{*}{\textbf{Model name}} & \multicolumn{2}{c}{\textbf{DynaHate}} & \multicolumn{2}{c}{\textbf{Hateeval}} \\
\cline{2-5} 
 & \textbf{Accuracy} & \textbf{F1} & \textbf{Accuracy} & \textbf{F1} \\
\hline
\shortstack{\textbf{TinyLlama-1.1B}} & 0.80 & 0.81 & 0.79 & 0.77 \\
\textbf{phi-2} & 0.80 & 0.83 & 0.79 & 0.78 \\
\textbf{opt-1.3b} & \textbf{0.82} & \textbf{0.83} & \textbf{0.80} & \textbf{0.81} \\
\hline
\end{tabular}
\end{table}

%%%%%%%%%%%%%%%%%%%%%%%%%%%%%%%%%%%%%%

The fine-tuning process has shown remarkable improvements across all models and methodologies compared to their respective base models.  It is observed from Table \ref{base_model}, \ref{lora} \& \ref{adapter}, initially, TinyLlama, although exhibiting moderate accuracy and F1 scores, underwent a substantial transformation post-fine-tuning.

The adapter-based fine-tuned models consistently displayed improvements, as shown in Table \ref{adapter}. For instance, TinyLlama saw its accuracy rise to 0.71 on Dynahate and 0.69 on Hateeval, with F1 scores reaching 0.75 and 0.70, respectively. Likewise, phi-2 achieved accuracies of 0.70 and 0.72 on Dynahate and Hateeval, respectively, alongside F1 scores of 0.76 and 0.71. Meanwhile, opt-1.3b attained accuracies of 0.71 on both datasets, with F1 scores of 0.71 on Dynahate and 0.74 on Hateeval. From Table \ref{adapter}, we note that phi2 achieves a higher F1 score with a slightly lower accuracy for the Dynahate dataset. However, for the Hateeval dataset, opt-1.3b exhibits higher accuracy and an F1 score improvement of 3-4\% compared to the other two models

With the LoRa technique (see, Table \ref{lora}), its accuracy surged from 0.50 to 0.80 on Dynahate and from 0.56 to 0.79 on Hateeval, accompanied by notable F1 score enhancements, rising from 0.61 to 0.81 and from 0.36 to 0.77, respectively. Similarly, fine-tuning with LoRa significantly improved phi-2, elevating its accuracy from 0.52 to 0.80 on Dynahate and from 0.22 to 0.79 on Hateeval, with F1 scores jumping from 0.66 to 0.83 and from 0.24 to 0.78, respectively. Opt-1.3b, another model subjected to fine-tuning using LoRa, witnessed impressive accuracy increments from 0.53 to 0.82 on Dynahate and from 0.47 to 0.77 on Hateeval, with F1 scores soaring from 0.54 to 0.83 and from 0.25 to 0.70, respectively. Analysis of Table \ref{lora} reveals that the opt-1.3b model demonstrates a 2\% increase in accuracy and a 1-2\% improvement in F1 score for the Dynahate dataset. Additionally, for the Hateeval dataset, its accuracy improves by 1\%, and the F1 score sees a boost of 3-4\%, when compared to the other two models.

% In general, the fine-tuning process significantly improved the performance of all models across both datasets. Notably, fine-tuning with the LoRa technique led to remarkable improvements in accuracy and F1 scores for all models. For instance, PY007/TinyLlama-1.1B-step-50K-105b, microsoft/phi-2, and facebook/opt-1.3b witnessed substantial accuracy increments and F1 score enhancements post-fine-tuning.
In general, fine-tuning, especially using the LoRa technique, significantly improved the performance of all models across both datasets. Notably, the opt-1.3b model consistently delivered strong performance, indicating its robustness in hate speech detection tasks. It is worth noting that opt-1.3b outperformed the larger phi2-2 model and also performed better than the slightly smaller model, tinyllama. Furthermore, adapter-based fine-tuned models,  also exhibited consistent improvements, suggesting the effectiveness of this approach in enhancing model performance.

% The fine-tuning process significantly improved the performance of all models across both datasets. Notably, fine-tuning with the LoRa technique led to remarkable improvements in accuracy and F1 scores for all models. For instance, PY007/TinyLlama-1.1B-step-50K-105b, microsoft/phi-2, and facebook/opt-1.3b witnessed substantial accuracy increments and F1 score enhancements post-fine-tuning. Furthermore, adopter-based fine-tuned models also exhibited consistent improvements, suggesting the effectiveness of this approach in enhancing model performance.

\section{Conclusion and Future Work}
This study pioneers the use of various tiny GPT-based tiny large language models (LLMs) for hate speech detection. We explore two different fine-tuning approaches and demonstrate that fine-tuned LLMs significantly outperform pre-trained models.
Overall, the results suggest that fine-tuning, particularly with the LoRa technique, is crucial for enhancing the performance of base models in hate speech detection tasks. Among the models evaluated, opt-1.3b consistently demonstrated strong performance across both datasets, indicating its robustness in this domain. Future work could focus on exploring additional fine-tuning techniques and conducting more extensive experiments to further improve the efficacy of hate speech detection models. Additionally, investigating the generalizability of these models across different languages and cultural contexts could be a promising direction for future research.

\bibliographystyle{IEEEtran}
\bibliography{custom}

\begin{thebibliography}{10}
\providecommand{\url}[1]{#1}
\csname url@samestyle\endcsname
\providecommand{\newblock}{\relax}
\providecommand{\bibinfo}[2]{#2}
\providecommand{\BIBentrySTDinterwordspacing}{\spaceskip=0pt\relax}
\providecommand{\BIBentryALTinterwordstretchfactor}{4}
\providecommand{\BIBentryALTinterwordspacing}{\spaceskip=\fontdimen2\font plus
\BIBentryALTinterwordstretchfactor\fontdimen3\font minus \fontdimen4\font\relax}
\providecommand{\BIBforeignlanguage}[2]{{%
\expandafter\ifx\csname l@#1\endcsname\relax
\typeout{** WARNING: IEEEtran.bst: No hyphenation pattern has been}%
\typeout{** loaded for the language `#1'. Using the pattern for}%
\typeout{** the default language instead.}%
\else
\language=\csname l@#1\endcsname
\fi
#2}}
\providecommand{\BIBdecl}{\relax}
\BIBdecl

\bibitem{davidson2017automated}
T.~Davidson, D.~Warmsley, M.~Macy, and I.~Weber, ``Automated hate speech detection and the problem of offensive language,'' in \emph{Proceedings of the international AAAI conference on web and social media}, vol.~11, no.~1, 2017, pp. 512--515.

\bibitem{fortuna2018survey}
P.~Fortuna and S.~Nunes, ``A survey on automatic detection of hate speech in text,'' \emph{ACM Computing Surveys (CSUR)}, vol.~51, no.~4, pp. 1--30, 2018.

\bibitem{schmidt2017survey}
A.~Schmidt and M.~Wiegand, ``A survey on hate speech detection using natural language processing,'' in \emph{Proceedings of the fifth international workshop on natural language processing for social media}, 2017, pp. 1--10.

\bibitem{malik2022deep}
J.~S. Malik, G.~Pang, and A.~v.~d. Hengel, ``Deep learning for hate speech detection: a comparative study,'' \emph{arXiv preprint arXiv:2202.09517}, 2022.

\bibitem{zhang2018detecting}
Z.~Zhang, D.~Robinson, and J.~Tepper, ``Detecting hate speech on twitter using a convolution-gru based deep neural network,'' in \emph{The Semantic Web: 15th International Conference, ESWC 2018, Heraklion, Crete, Greece, June 3--7, 2018, Proceedings 15}.\hskip 1em plus 0.5em minus 0.4em\relax Springer, 2018, pp. 745--760.

\bibitem{badjatiya2017deep}
P.~Badjatiya, S.~Gupta, M.~Gupta, and V.~Varma, ``Deep learning for hate speech detection in tweets,'' in \emph{Proceedings of the 26th international conference on World Wide Web companion}, 2017, pp. 759--760.

\bibitem{vaswani2017attention}
A.~Vaswani, N.~Shazeer, N.~Parmar, J.~Uszkoreit, L.~Jones, A.~N. Gomez, {\L}.~Kaiser, and I.~Polosukhin, ``Attention is all you need,'' \emph{Advances in neural information processing systems}, vol.~30, 2017.

\bibitem{devlin2018bert}
J.~Devlin, M.-W. Chang, K.~Lee, and K.~Toutanova, ``Bert: Pre-training of deep bidirectional transformers for language understanding,'' \emph{arXiv preprint arXiv:1810.04805}, 2018.

\bibitem{clark2020electra}
K.~Clark, M.-T. Luong, Q.~V. Le, and C.~D. Manning, ``Electra: Pre-training text encoders as discriminators rather than generators,'' \emph{arXiv preprint arXiv:2003.10555}, 2020.

\bibitem{lewis2019bart}
M.~Lewis, Y.~Liu, N.~Goyal, M.~Ghazvininejad, A.~Mohamed, O.~Levy, V.~Stoyanov, and L.~Zettlemoyer, ``Bart: Denoising sequence-to-sequence pre-training for natural language generation, translation, and comprehension,'' \emph{arXiv preprint arXiv:1910.13461}, 2019.

\bibitem{mozafari2020bert}
M.~Mozafari, R.~Farahbakhsh, and N.~Crespi, ``A bert-based transfer learning approach for hate speech detection in online social media,'' in \emph{Complex Networks and Their Applications VIII: Volume 1 Proceedings of the Eighth International Conference on Complex Networks and Their Applications COMPLEX NETWORKS 2019 8}.\hskip 1em plus 0.5em minus 0.4em\relax Springer, 2020, pp. 928--940.

\bibitem{aluru2021deep}
S.~S. Aluru, B.~Mathew, P.~Saha, and A.~Mukherjee, ``A deep dive into multilingual hate speech classification,'' in \emph{Machine Learning and Knowledge Discovery in Databases. Applied Data Science and Demo Track: European Conference, ECML PKDD 2020, Ghent, Belgium, September 14--18, 2020, Proceedings, Part V}.\hskip 1em plus 0.5em minus 0.4em\relax Springer, 2021, pp. 423--439.

\bibitem{maity2022mtbullygnn}
K.~Maity, T.~Sen, S.~Saha, and P.~Bhattacharyya, ``Mtbullygnn: a graph neural network-based multitask framework for cyberbullying detection,'' \emph{IEEE Transactions on Computational Social Systems}, 2022.

\bibitem{awal2023model}
M.~R. Awal, R.~K.-W. Lee, E.~Tanwar, T.~Garg, and T.~Chakraborty, ``Model-agnostic meta-learning for multilingual hate speech detection,'' \emph{IEEE Transactions on Computational Social Systems}, 2023.

\bibitem{guo2023investigation}
K.~Guo, A.~Hu, J.~Mu, Z.~Shi, Z.~Zhao, N.~Vishwamitra, and H.~Hu, ``An investigation of large language models for real-world hate speech detection,'' in \emph{2023 International Conference on Machine Learning and Applications (ICMLA)}.\hskip 1em plus 0.5em minus 0.4em\relax IEEE, 2023, pp. 1568--1573.

\bibitem{pendzel2023generative}
S.~Pendzel, T.~Wullach, A.~Adler, and E.~Minkov, ``Generative ai for hate speech detection: Evaluation and findings,'' \emph{arXiv preprint arXiv:2311.09993}, 2023.

\bibitem{vidgen2020learning}
B.~Vidgen, T.~Thrush, Z.~Waseem, and D.~Kiela, ``Learning from the worst: Dynamically generated datasets to improve online hate detection,'' \emph{arXiv preprint arXiv:2012.15761}, 2020.

\bibitem{basile2019semeval}
V.~Basile, C.~Bosco, E.~Fersini, D.~Nozza, V.~Patti, F.~M.~R. Pardo, P.~Rosso, and M.~Sanguinetti, ``Semeval-2019 task 5: Multilingual detection of hate speech against immigrants and women in twitter,'' in \emph{Proceedings of the 13th international workshop on semantic evaluation}, 2019, pp. 54--63.

\bibitem{zhang2024tinyllama}
P.~Zhang, G.~Zeng, T.~Wang, and W.~Lu, ``Tinyllama: An open-source small language model,'' \emph{arXiv preprint arXiv:2401.02385}, 2024.

\bibitem{li2023textbooks}
Y.~Li, S.~Bubeck, R.~Eldan, A.~Del~Giorno, S.~Gunasekar, and Y.~T. Lee, ``Textbooks are all you need ii: phi-1.5 technical report,'' \emph{arXiv preprint arXiv:2309.05463}, 2023.

\bibitem{zhang2023opt}
S.~Zhang, S.~Roller, N.~Goyal, M.~Artetxe, M.~Chen, S.~Chen, C.~Dewan, M.~Diab, X.~Li, X.~V. Lin \emph{et~al.}, ``Opt: Open pre-trained transformer language models, 2022,'' \emph{URL https://arxiv. org/abs/2205.01068}, vol.~3, pp. 19--0, 2023.

\bibitem{hu2021LoRA}
E.~J. Hu, Y.~Shen, P.~Wallis, Z.~Allen-Zhu, Y.~Li, S.~Wang, L.~Wang, and W.~Chen, ``Lora: Low-rank adaptation of large language models,'' \emph{arXiv preprint arXiv:2106.09685}, 2021.

\bibitem{houlsby2019parameter}
N.~Houlsby, A.~Giurgiu, S.~Jastrzebski, B.~Morrone, Q.~De~Laroussilhe, A.~Gesmundo, M.~Attariyan, and S.~Gelly, ``Parameter-efficient transfer learning for nlp,'' in \emph{International conference on machine learning}.\hskip 1em plus 0.5em minus 0.4em\relax PMLR, 2019, pp. 2790--2799.

\end{thebibliography}

\begin{IEEEbiography}[{\includegraphics[width=1in,height=1.05in,clip]{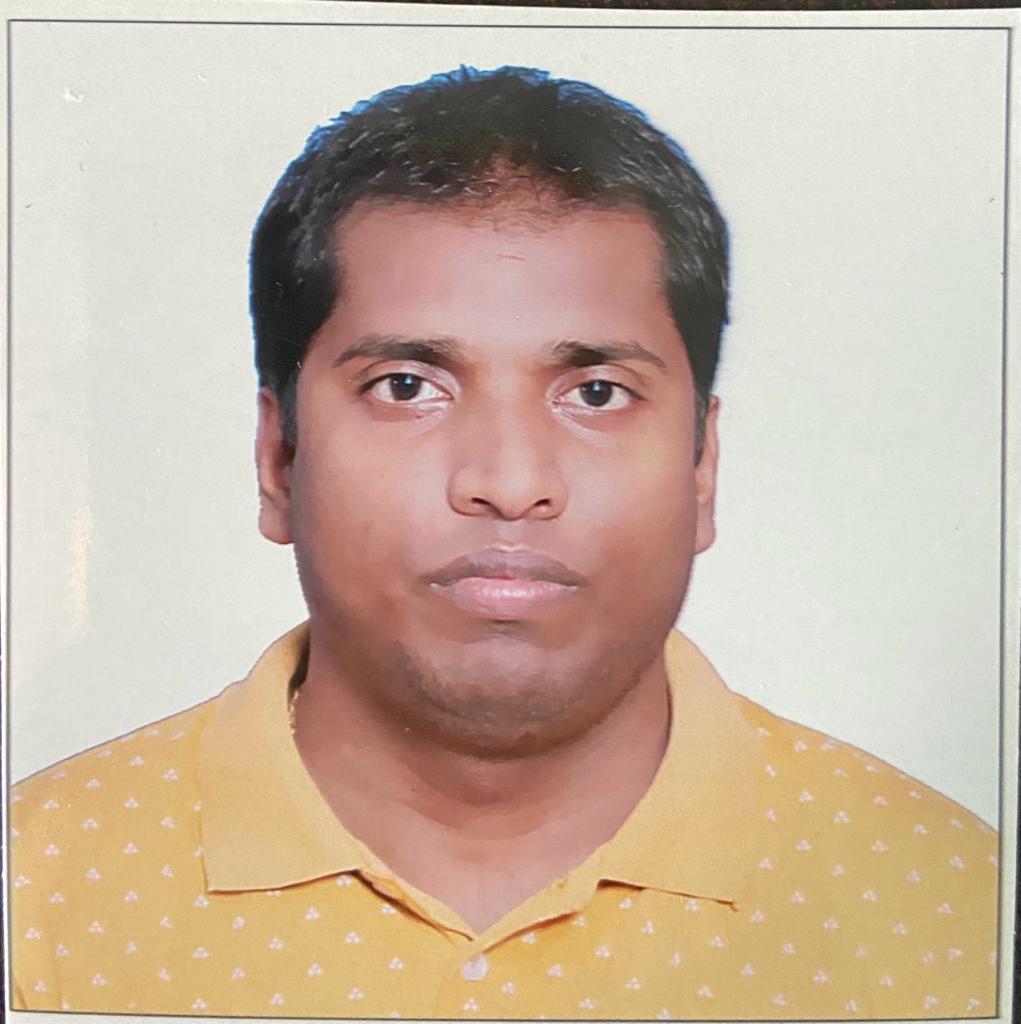}}]{Tanmay Sen} is a seasoned Lead Data Scientist at Ericsson, Kolkata. He earned his B.Sc. (Hons) and M.Sc. degrees in Mathematics from the University of Calcutta in 2009 and 2011, respectively. He pursued his M.Tech in Mathematics and Computing and later obtained his PhD in Statistics from the Indian Institute of Technology Patna in 2014 and 2019, respectively. His research spans various cutting-edge domains, including Deep Learning, Federated Learning, Meta Learning, Graph Neural Networks, NLP, Time series and Survival Analysis.
\end{IEEEbiography}

\begin{IEEEbiography}[{\includegraphics[width=1in,height=1.05in,clip]{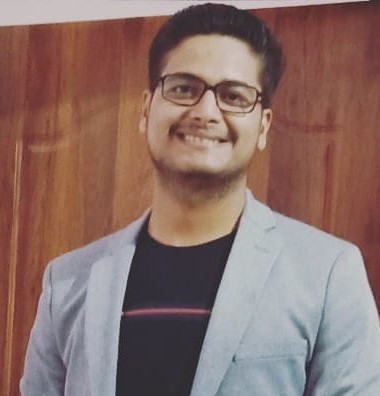}}]
{Ansuman Das} is an experienced Lead Data Scientist with approximately 10 years of expertise in data science, analytics, and machine learning. He has held significant positions at prominent companies, presently serving as a Lead Data Scientist at Ericsson since 2021. He completed his B.Tech in Computer Science $\&$ Engineering from IIIT Bhubaneswar and pursued his M.Tech in Software Systems from BITS Pilani. Currently, he is enrolled in an MA program in Economics at IGNOU. His research interests include Deep Learning, Generative AI, Natural Language Processing, and Cloud Computing.
\end{IEEEbiography}

\begin{IEEEbiography}[{\includegraphics[width=1in,height=1.05in,clip]{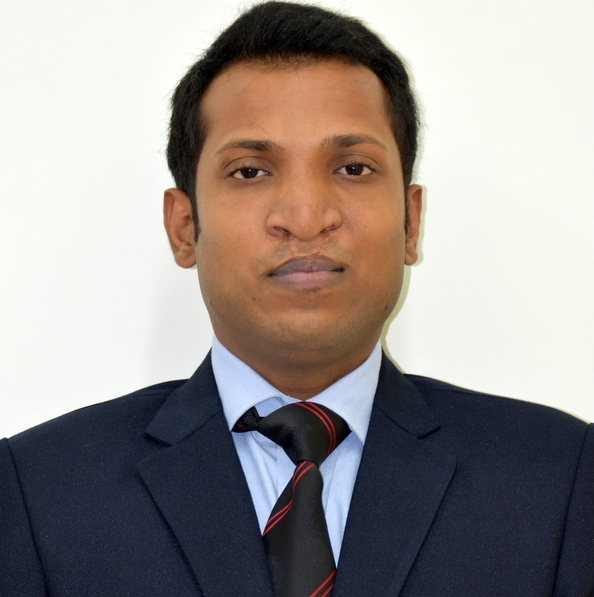}}]{Mrinmay Sen} is a joint research scholar in the Department of Artificial Intelligence, Indian Institute of Technology Hyderabad and the Department of Computing Technologies, Swinburne University of Technology, Australia. He completed his M.Tech from Indian Institute of Technology Dhanbad and B.E. from Jadavpur University, Kolkata. His research encompasses Federated Optimization and its applications in real-life scenarios, as well as Computer Vision and deep learning.
\end{IEEEbiography}

% \addvspace{-2.3cm}

\end{document}